\pdfoutput=1

\documentclass[11pt]{article}

\usepackage{acl}
\usepackage{acl}

\usepackage{times}
\usepackage{latexsym}
\usepackage{booktabs} 
\usepackage{import}
\usepackage{graphicx}
\usepackage{caption}
\usepackage{subcaption}
\usepackage[T1]{fontenc}

\usepackage[utf8]{inputenc}

\usepackage{microtype}

\title{Training Models to Generate, Recognize, and Reframe Unhelpful Thoughts}

  \author{Mounica Maddela\thanks{* Equal contribution.}\\
  Georgia Tech \& \\
  Meta AI
  \And
  Megan Ung$^*$\\
  Meta AI 
  \And
  Jing Xu\\
 Meta AI 
  \AND
 Andrea Madotto \\
  Meta AI 
  \And
 Heather Foran\\
Klagenfurt University
 \And
  Y-Lan Boureau \\
  Meta AI 
}

\begin{document}
\maketitle
\begin{abstract}
Many cognitive approaches to well-being, such as recognizing and reframing unhelpful thoughts, have received considerable empirical support over the past decades, yet still lack truly widespread adoption in self-help format. A barrier to that adoption is a lack of adequately specific and diverse dedicated practice material. This work examines whether current language models can be leveraged to  both produce a virtually unlimited quantity of practice material illustrating standard unhelpful thought patterns matching specific given contexts, and generate suitable positive reframing proposals. We propose \textsc{PatternReframe}, a novel dataset of about 10k examples of thoughts containing unhelpful thought patterns conditioned on a given persona, accompanied by about 27k positive reframes. By using this dataset to train and/or evaluate current models, we show that existing models can already be powerful tools to help generate an abundance of tailored practice material and hypotheses, with no or minimal additional model training required.
\end{abstract}

\section{Introduction}

Cognitive Behavioral Therapy (CBT) \cite{beck1963, beck1976} is one of the most robustly validated approaches in psychology \cite{cbteffects2012, david2018}. A core pillar of CBT consists in identifying and reframing unhelpful ways of thinking. Low-intensity CBT interventions have shown promise in self-help formats \cite{shafran2021, williams2001}, yet a lack of sufficient practice material suited to people's specific circumstances is a barrier to adoption \citep{helgadottir2009online}.

Through prompting, control tokens, or adequate conditioning, modern language models can guide generation of language towards desired outcomes, such as conforming to a given persona \citep{zhang-etal-2018-personalizing}, style \citep{ziems-etal-2022-inducing}, or level of confidence \citep{mielke2022reducing}. This makes them a potentially powerful practice aid for learning cognitive reframing techniques.
A major barrier  
is the lack of publicly available data. Most 
existing work in 
natural language processing (NLP) for CBT focuses on interactions between patients and mental health professionals, which are 
not publicly available \cite{mieskes-stiegelmayr-2018-preparing, rojas-barahona-etal-2018-deep, shreevastava-foltz-2021-detecting}. \citet{ziems-etal-2022-inducing} released the first public dataset for reframing tweets 
marked with a hashtag indicating stress, using known reframing techniques, but it does not specifically look at the categories of unhelpful thinking used in CBT, and uses existing tweets rather than allowing the generation of examples suited to a particular situation.

In this work, we propose\footnote{The dataset and task have been released through the ParlAI framework \citep{miller2017parlai} and are available at \url{https://github.com/facebookresearch/ParlAI/tree/main/projects/reframe_thoughts}} a novel dataset, \textsc{PatternReframe}, consisting in $\sim$10k crowdsourced examples of thoughts containing ten classical types of unhelpful thought patterns \citep{burns1980feeling}, conditioned on personas, matched with crowdsourced proposals of reframing that do not exhibit the patterns.
We introduce two controllable text-to-text generation tasks on the dataset:  (1) generating and (2) reframing unhelpful thoughts, given a persona and pattern as the context. We also define a classification task to identify the unhelpful thought pattern, given a persona and a thought. We train and evaluate different fine-tuned and few-shot approaches for the tasks, and show that these approaches perform reasonably well on the tasks. 

\section{Related Work}

\subsection{NLP for Mental Health}
 Recent work has used linguistic features and pretrained language models to identify mental health conditions such as anxiety \cite{owen-etal-2020-towards, shreevastava-foltz-2021-detecting, fine-etal-2020-assessing}, depression \cite{wolohan-etal-2018-detecting, poswiata-perelkiewicz-2022-opi, ji-etal-2022-mentalbert}, schizophrenia \cite{jiang-etal-2020-detection, mitchell-etal-2015-quantifying, sarioglu-kayi-etal-2017-predictive}, and post-traumatic stress disorder \cite{coppersmith-etal-2015-clpsych}. Most of these works annotate social media posts to create datasets for the task, and then train and evaluate different classification models.  \citet{shreevastava-foltz-2021-detecting} and \citet{rojas-barahona-etal-2018-deep} created datasets for identifying unhelpful thoughts by annotating patient-therapist interactions and finetuned different pretrained models for the task. However, these datasets are not publicly available.
 
 The closest work to ours is that of \citet{ziems-etal-2022-inducing}, which introduces a reframing task, releases a parallel corpus of reframed sentences, and uses controllable text generation models to reframe social media content from Twitter that was marked as expressing stress. However, the source social media material is not conditioned on personas, or focused on the classical unhelpful thought patterns from CBT. Our work introduces conditioning on personas and classical unhelpful thought patterns, and extends the reframing task to  identifying and generating thoughts matching a given persona and unhelpful pattern. 

\subsection{Controllable Text Generation}

Controllable text generation approaches using pretrained language models (PLMs) typically fall into four categories: (i) prompt-based methods that either construct templates for PLMs to complete \cite{jiang-etal-2020-know, schick-schutze-2021-exploiting, schick-schutze-2021-shot} or finetune a task-specific layer to guide the generation \cite{li-liang-2021-prefix, lester-etal-2021-power}, (ii) finetuning methods that either use labelled data prepended with controlled attributes \cite{ziems-etal-2022-inducing, fan-etal-2018-hierarchical, martin-etal-2020-controllable,ross2022tailor} or define a task-specific reward function using reinforcement learning \cite{https://doi.org/10.48550/arxiv.1909.08593, liu-etal-2020-data},  (iii) post-processing methods that train discriminator models to guide the generation towards a specific criterion during decoding \cite{https://doi.org/10.48550/arxiv.1912.02164, hua-wang-2020-pair, xu-etal-2020-megatron}, and (iv) pretraining methods that pre-train PLMs from the start with different control tokens prepended to the input \cite{https://doi.org/10.48550/arxiv.1909.05858}. In our work, we experiment with prompt-based and finetuning methods.

\section{Identifying and Reframing Unhelpful Thoughts}
\label{sec:framework}
\setlength{\tabcolsep}{4pt}
\begin{table*}[ht!]
\centering
\small
\begin{tabular}{p{5.1cm}|p{10.0cm}}
\toprule
\textbf{Unhelfpul Thought Patterns and their distribution} & \textbf{Example Thoughts and their Rewrites  that remove the pattern}  \\ \midrule
\textbf{Catastrophizing} by giving greater weight to the worst possible outcome. 
\newline (1024 thoughts / 2826 rewrites) & 
\textit{My mom hasnt come home from work yet. I hope the store isn`t getting robbed!} 
\newline \textbf{Rewrite:} \textit{My mom hasn't come home from work yet. She must have gotten swamped. I'll cook dinner now so it's ready when she gets home.} \\ 
\midrule
\textbf{Discounting the positive:} experiences by insisting that they ``don't count''. 
\newline (970 thoughts / 2680 rewrites)
& \textit{My restaurant is the most popular in my city, but that's just luck.} 
\newline \textbf{Rewrite:} \textit{My restaurant is the most popular in the city. I suppose all my hard work has paid off.} \\
\midrule 
\textbf{Overgeneralization} is making faulty generalizations from insufficient evidence. 
\newline (983 thoughts / 2747 rewrites)
&  \textit{My nephews didn't want to spend the weekend with me this week. I must not be as good of an aunt as I thought.} 
\newline \textbf{Rewrite}: \textit{My nephews didn't want to spend the weekend with me this week. They must be busy.}  \\
\midrule
\textbf{Personalization} is assigning a disproportionate amount of personal blame to oneself. (934 thoughts / 2544 rewrites) &
\textit{My sister was not happy with the makeup look I did for her. I am a bad artist.}
\newline \textbf{Rewrite}: \textit{My sister was not happy with the makeup I did for her, next time I'll try something different.}
\\
\midrule
\textbf{All-or-nothing} is viewing things as either good or bad and nothing in-between. \newline (952 thoughts / 2628 rewrites) &
\textit{The school christmas choir concert got canceled. This holdiday season is ruined.} 
\newline \textbf{Rewrite}: \textit{Even though the choir concert got canceled there are still other fun activities to do on the holiday.} \\
\midrule
\textbf{Mental Filtering} occurs when an individual dwells only on the negative details of a situation.
\newline (936 thoughts / 2562 rewrites) &
\textit{It's nice to enjoy the sea breeze when you live near the ocean but it's not worth it when you think of all the sand getting dragged into your home and all the tourists making so much noise at the beach.} 
\newline \textbf{Rewrite}: \textit{I am so fortunate to live where I can enjoy the sea breeze.  Not everyone is this lucky.}  \\
\midrule
\textbf{Mind Reading} is inferring a person`s probable (usually negative) thoughts from their behavior. 
\newline (992 thoughts / 2688 rewrites)& 
\textit{I auditioned for the surf team and the coach avoided me. I am sure it is because he does not like my skills.} 
\newline \textbf{Rewrite}: \textit{I auditioned for the surf team and the coach avoided me. I'm sure the coach always tries to appear neutral during try-outs.}  \\
\midrule
\textbf{Fortune Telling} is predicting outcomes (usually negative) of events. \newline (997 thoughts / 2758 rewrites) & 
\textit{I didn't make it to Yellowstone this year, I am never going to go to that park.}
\newline \textbf{Rewrite}: \textit{I didn't get to go to Yellowstone this year, I will work extra hard and save up to definitely go next year!} \\
\midrule
\textbf{Should statements}, where a person demands particular behaviors regardless of the realistic circumstances. \newline (921 thoughts / 2413 rewrites) & 
\textit{I prefer texting over phone calls. People should never call me and expect me to answer.}
\newline \textbf{Rewrite}: \textit{Just because I like texting doesn't mean everyone needs to like it.} \\
\midrule 
\textbf{Labeling and mislabeling} is attributing a person's actions to their character rather than the situation. \newline (960 thoughts / 2661 rewrites) & 
\textit{I fell off my skateboard yesterday, I'm a terrible athlete.}
\newline \textbf{Rewrite}: \textit{I fell off my skateboard yesterday, but even the best crash sometimes.} 
\\
\bottomrule
\end{tabular}
\setlength{\belowcaptionskip}{-8pt}
\caption{Examples of unhelpful thoughts and their reframed versions from our \textsc{PatternReframe} dataset. The thought pattern definitions are derived from Wikipedia.}
\label{table:examples_from_data}
\end{table*}

We use the ten categories of unhelpful thought patterns described in lay terms in a widely used CBT self-help book used for bibliotherapy  \citep{burns1980feeling}. Table \ref{table:examples_from_data} lists these categories and provides examples for each category. For reframing unhelpful thoughts, we follow \citet{ziems-etal-2022-inducing}, who describe five reframing strategies  based on positive psychology \cite{pospsy2007}: (i) \textit{Growth Mindset:} Focusing on learning from challenges and improving the skills needed to deal with a difficult situation; (ii) \textit{Optimism:} Directing the attention towards the positive aspects of the situation and expressing gratitude while still acknowledging the negative aspects; (iii) \textit{Impermanence:} Understanding that adversities are inevitable and temporary and focusing on accepting the situation; (iv) \textit{Neutralizing:} Challenging unhelpful thoughts that are far from reality and replacing them with realistic neutral alternatives; (v) \textit{Self-affirmation:} Reflecting on core values to ground oneself in a difficult situation. Note that other reframing strategies exist, such as ``\textit{being mindful}'' \cite{robertson2012build}, or  ``\textit{focusing on forgiveness and compassion}'' \cite{cbt_comp2010}. 
We provide the above five strategies only as a starting point, but crowd workers are free to use other strategies.

\section{\textsc{PatternReframe} Dataset}

\subsection {Data Collection}
\label{sec:annotation_pipeline}
We briefly explain the four-step data collection process used to crowdsource the dataset. We provide further data collection details and snapshots of the interface in Appendix \ref{app:annDetails} and \ref{app:ann_interface}.

\subsubsection {Task 1: Writing Unhelpful Thoughts}
\label{sec:step1}
In order to generate unhelpful thoughts that match a diversity of contexts and situations, we use 
personas from the \textsc{Persona-Chat} dataset \cite{zhang-etal-2018-personalizing} as context for writing unhelpful thoughts.  
We give a persona and one of the ten unhelpful thought patterns to the crowdsource workers, and ask them to write sentences that both are consistent with the given persona, and exhibit the given unhelpful thought pattern. 

\subsubsection {Task 2: Categorizing Unhelpful Thoughts}
\label{sec:step2}
Unhelpful thoughts can exhibit multiple patterns, and the patterns themselves are overlapping rather than distinct \cite{burns1980feeling}. 
In order to capture this, as well as filter out low-quality crowdsourced data,
we use a second crowdsourcing task requesting workers to label the previously generated thoughts.
Workers are given a thought and 
the list of unhelpful patterns, and select all the patterns that appear in the thought.
The annotators can choose a ``None'' option in case the thought is irrelevant or nonsensical. We collect five annotations for each thought, and discard the thoughts that are marked ``None'' by a majority of annotators. 

\subsubsection {Task 3: Reframing Unhelpful Thoughts}
\label{sec:step3}
In a third task, we ask crowdworkers to rewrite thoughts containing unhelpful patterns, in a more helpful way, similar to the task in \citet{ziems-etal-2022-inducing}.
We give crowdworkers a thought and the persona and unhelpful pattern that were used to generate it, and ask them to rewrite the thought in a way that still aligns with the context, but does not contain the unhelpful pattern. We also show the five reframing strategies described in \S \ref{sec:framework} to aid the workers in reframing the thoughts, and ask them to select what strategy they used, if any.
Note that the strategies are only provided as suggestions, and the workers are free to 
reframe the thought in other appropriate ways. We collect three rewrites for each thought. 

\subsubsection {Task 4: Evaluating the Rewrites of Unhelpful Thoughts}
\label{sec:step4}
Finally, we assess the quality of the rewrites as follows: 
workers are given a persona, unhelpful thought pattern, generated thought, along with three rewrites. They are asked to select which rewrites successfully remove the unhelpful pattern while not logically contradicting the source (following \citet{ziems-etal-2022-inducing}). If worker selects a valid rewrite, we further ask them to identify which of the five proposed reframing strategies were used, if any. We collect five annotations for each set, and include only the rewrites that are marked as ``valid'' by a majority of annotators.

\subsection {Data Quality}

We use the Mephisto\footnote{https://github.com/facebookresearch/Mephisto} and Amazon Mechanical Turk\footnote{Our crowdsourcing tasks pay workers well above minimum wage.} platforms to collect crowdsource data. 
We use the labeling tasks (2nd and 4th task) to select a pool of high-quality workers (that is, crowdsource workers whose generative work was validated by a majority of separate annotators in a separate labeling task), after first seeding the set of annotators through manual inspection of a first batch of data.
We use only selected annotators for evaluation tasks (tasks 2 and 4). We first kept the generative text tasks (tasks 1 and 3) open to all workers. We expanded the list of selected workers after every iteration by adding new workers that had completed at least five generative text tasks with at least 80\% of generated text validated through the evaluation tasks. We ended up with 524 qualified workers after nine rounds of the entire pipeline, where each iteration started with a batch of 500 thoughts. Once we gathered $> 500$ qualified workers, we restricted all the tasks to the selected pool. In the final dataset, we included only the annotations provided by these selected workers. 

Along with the selected pool of workers, we also included onboarding tasks (details in \S \ref{app:annDetails}) to ensure that the workers adequately understood the concept of reframing thoughts. Only the workers who passed the onboarding tasks were qualified to work on the actual tasks. We calculated inter-annotator agreement using Krippendorf’s Alpha, which was 0.355 for the second task and 0.454 for the fourth task.\footnote{We compute  Krippendorf’s Alpha for the binary pattern-level judgments from the the second task and the binary reframe-level judgements from the fourth task.}

\subsection {Data Analysis}

\subsubsection{Dataset Statistics}
\label{sec:dataset_stats}
\textsc{PatternReframe} contains 9,688 thoughts and 26,507 reframed versions of thoughts. We split the dataset into training, validation, and test sets of respective sizes 1,920 / 961 / 6,807 for thoughts, and 5,249 / 2,623 / 18,635 for reframed thoughts.
One thought can have up to three reframed versions, with an average of 2.74 rewrites / thought after filtering out lower-quality rewrites. The average word lengths of thoughts and rewrites are 19.1 and 23.9, respectively. 

\begin{figure}[t!]
  \centering
  \includegraphics[height=6.2cm, width=1.0\linewidth]{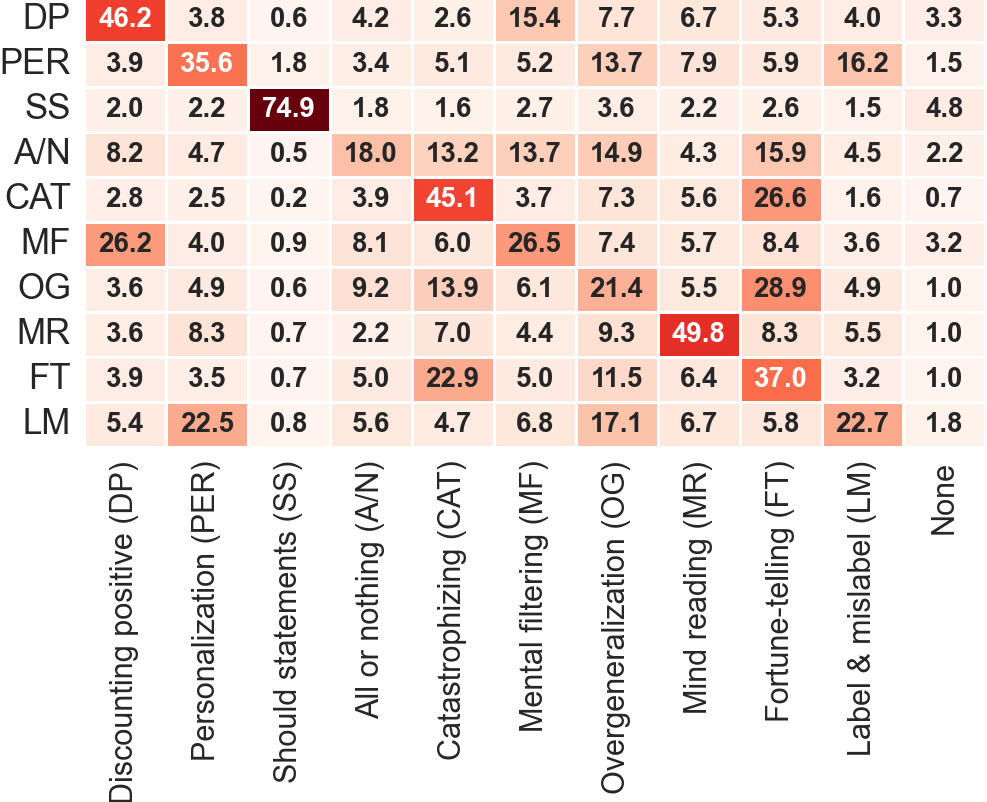}  
  \caption{Confusion matrix representing the distribution of unhelpful thoughts across different patterns in our dataset. Rows represent the patterns used to collect the unhelpful thoughts in the first task (\S \ref{sec:step1}). Columns represents the patterns chosen by annotators in the second task (\ref{sec:step2}). As expected, some related patterns such as Discounting the positive (DP) and Mental filtering (MF) exhibit strong cross-labeling.}
  \label{fig:category_vs_category}
\end{figure}

\subsubsection{Analysis of Unhelpful Thought Patterns}

Figure \ref{fig:category_vs_category} shows the distribution of thoughts across different patterns in our dataset, with initial conditioning pattern (1st task) in rows and annotator identified patterns (2nd task) in columns. 
As expected, there is a high overlap among some related patterns, e.g., \textit{Discounting the positive} /  \textit{Mental Filtering},  \textit{Fortune Telling}/ \textit{Catastrophizing}, and \textit{Personalization} / \textit{Labeling and Mislabeling}. \textit{All or Nothing Thinking} is difficult to distinguish, and shows high overlap with many categories. \textit{Mind Reading} and \textit{Should Statement} show the lowest amounts of overlap with other patterns.

\subsubsection{Analysis of Reframing Strategies:}

\begin{figure}[th!]
  \centering
  \includegraphics[width=1.0\linewidth]{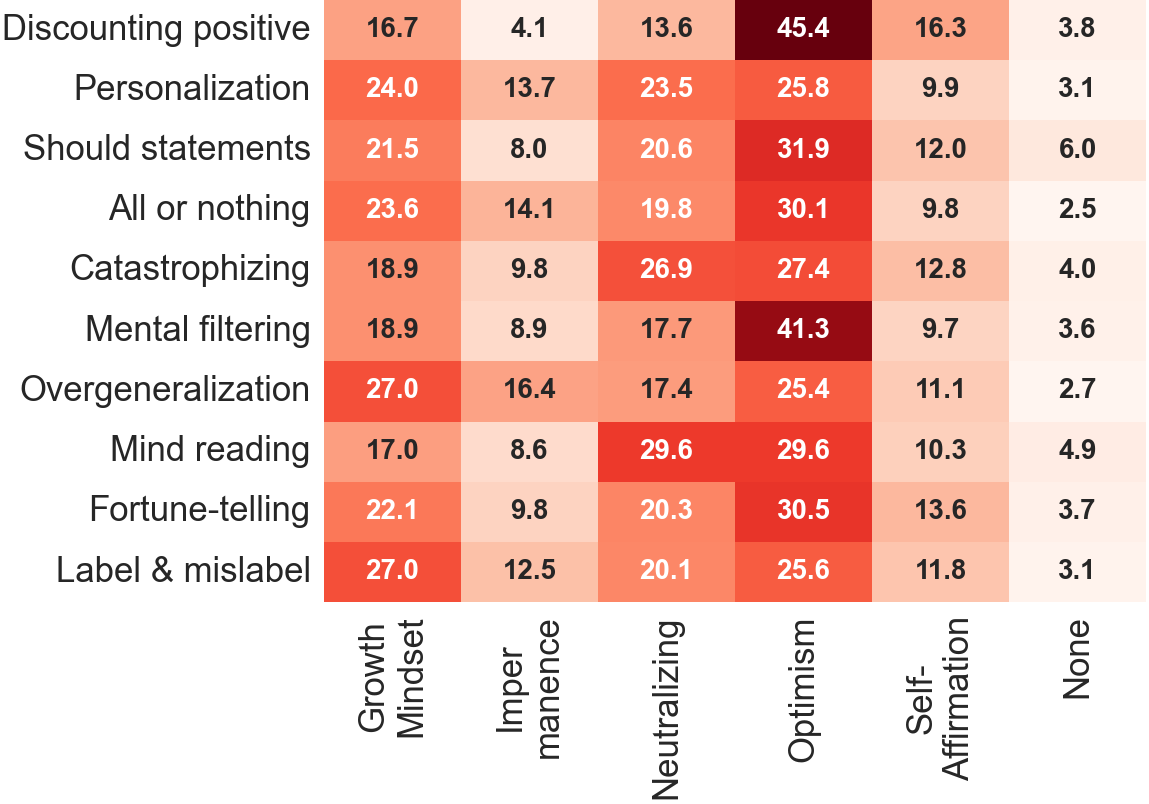}  
  \caption{Matrix showing the distribution of reframing strategies across different unhelpful thought patterns. Rows represent the unhelpful thought patterns and columns represent the reframing strategies.}
  \label{fig:category_vs_strategy}
\end{figure}

Figure \ref{fig:category_vs_strategy} shows the distribution of reframing strategies used to reframe the unhelpful thoughts in our dataset, among the five strategies proposed by \citet{ziems-etal-2022-inducing}. Here, we use the strategies identified by the workers in the fourth task of evaluating reframed thoughts.
Most rewritten thoughts make use of one of the five strategies, with very few being labeled as "None."
\textit{Growth Mindset} and \textit{Optimism} are the most commonly used reframing strategies, followed by \textit{Neutralizing} and \textit{Self-Affirmation}. \textit{Optimism} is especially common for patterns that focus on the negative aspects of the situation such as \textit{Discounting the positive} and \textit{Mental Filtering}.

\section{Models to Generate, Recognize, and Reframe Unhelpful Thoughts}
We train and evaluate different models using our \textsc{PatternReframe} dataset on three tasks: generating, identifying, and reframing unhelpful thoughts -- all conditioned on a given persona.

\subsection {Generating Unhelpful Thoughts}

\setlength{\tabcolsep}{4pt}
\begin{table*}[ht!]
\small
\centering
\begin{tabular}{l|ccccc|ccccc}
\toprule
& \multicolumn{5}{c|}{\textbf{Generating Unhelpful Thoughts}} 
& \multicolumn{5}{c}{\textbf{ Reframing Unhelpful Thoughts}}  \\
& \textbf{BLEU} & \textbf{ROUGE} & \textbf{BScore}  & \textbf{Dist-1} &  \textbf{Dist-2} 
& \textbf{BLEU} & \textbf{ROUGE} & \textbf{BScore}  & \textbf{Dist-1} &  \textbf{Dist-2} \\ \midrule 
BART 
& 25.3 & 23.9 & 89.0 & 0.021 & 0.087 
& 69.7 & 53.1 & 93.5 & 0.034 & 0.223
\\
T5
& 24.5 & \textbf{24.3} & 89.1 & 0.019 & 0.08
& 69.9 & \textbf{55.5} & 93.6 & 0.039 & 0.261 
\\
R2C2
& \textbf{25.5} & 24.1 & \textbf{89.2} & 0.023 & 0.1  
& \textbf{70.0} & 55.0 & \textbf{93.7} & 0.036 & 0.235   
\\
GPT3.5$\dagger$ 
& 24.9 & 19.2 & 88.1 & \textbf{0.196} & \textbf{0.586} 
&  51.5 & 41.2 & 91.7 & \textbf{0.204} & \textbf{0.633} 
\\
\midrule
Reference
& 100.0 & 100.0 & 100.0 & 0.044 & 0.304  
& 100.0 & 100.0 & 100.0 & 0.041 & 0.309 
\\
\bottomrule
\end{tabular}
\caption{Automatic evaluation results on the \textsc{PatternReframe} test set. We report \textbf{BLEU}, \textbf{ROUGE}, BERTScore (\textbf{BScore}), Distinct-1 (\textbf{Dist-1}), and Distinct-2 (\textbf{Dist-2}) metrics. $\dagger$We calculate metrics over 100 random generations because of our limited access to the GPT3.5 
 API (text-davinci-002).}
\label{table:reframe_generate_results}
\end{table*}

\subsubsection{Task and Data}
\label{sec:generating_task}
Given a persona and an unhelpful thought pattern, the goal is to generate a thought that exhibits the given pattern and aligns with the persona.  We formulate the task as a standard conditioned generation problem and optimize the maximum likelihood loss during training. We use the train, validation, and test splits described in \S \ref{sec:dataset_stats}.

\subsubsection{Methods}
\label{sec:generation_methods}
We evaluate methods based on fine-tuning and few-shot learning. We fine-tune BART-large \cite{lewis-etal-2020-bart}, T5-large \cite{2020t5}, and R2C2-3B \cite{sludge2022} (a BART-based language model specialized in dialogues). For the input, we concatenate the persona and the unhelpful thought pattern texts using a special delimiter token. We also generate responses with GPT3.5 \cite{instructgpt2022}, a state-of-the-art language model trained to follow human instructions, as a 1-shot method. We  generated thoughts for only 100 random inputs in the \textsc{PatternReframe} test set, since we had limited access to the API\footnote{https://openai.com/api/} to GPT3.5 (text-davinci-002)\footnote{In our experiments, we used text-davinci-002, since text-davinci-003 had not been released yet.}. We provide implementation details and examples of input prompts in Appendix \ref{app:implementation details} and \ref{app:prompt_examples}, respectively.

\begin{figure*}[th!]
    \centering
    \includegraphics[width=0.7\linewidth]{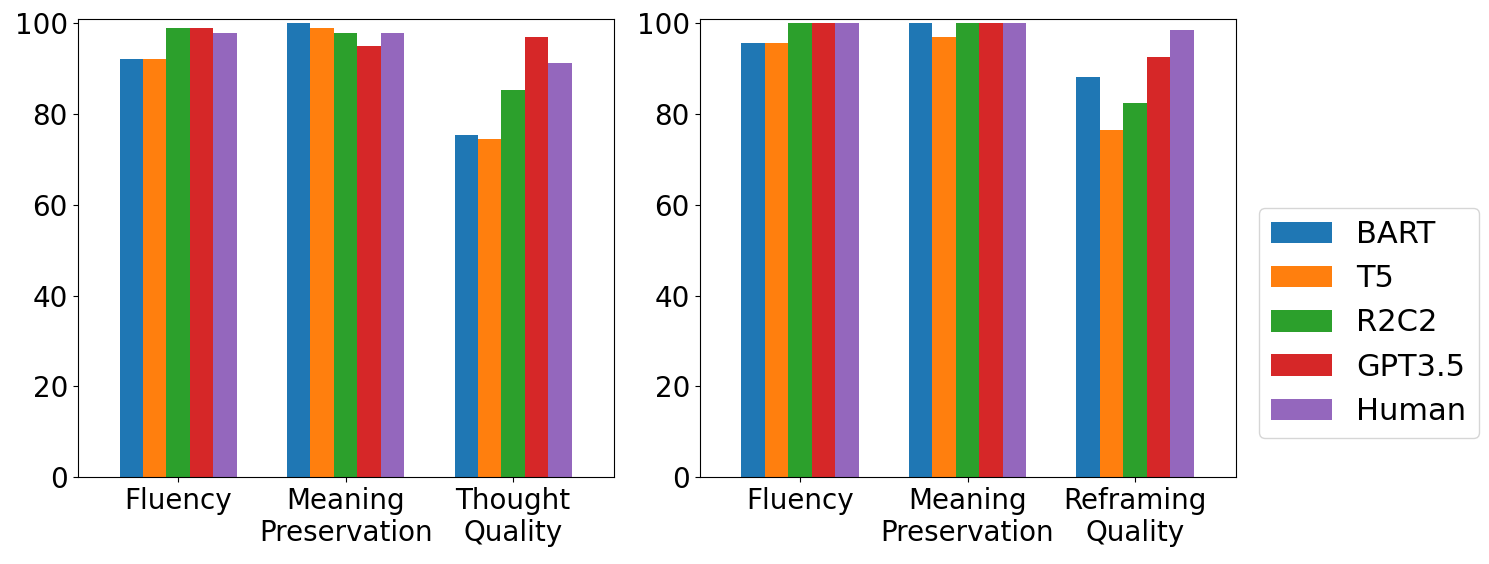}
    \caption{Human evaluation results for the tasks of generating (left) and reframing  (right) unhelpful thoughts. Y-axis shows the percentage of outputs rated positively by at least five of the nine annotators.}
    \label{fig:human_eval}
\end{figure*}

\subsubsection{Automatic Evaluation}
\label{sec:reframing_automatic_metrics}

Following previous work on text reframing \cite{ziems-etal-2022-inducing, chen-etal-2021-controlled-neural}, we report BLEU \cite{Papineni:2002:BMA:1073083.1073135}, ROUGE \cite{lin-2004-rouge}, and BERTScore \cite{zhang2020bertscore}, which capture the semantic similarity between the generated thought and the human reference. We also report distinct-1, and distinct-2 metrics to measure the diversity of the generations. Distinct-n \cite{li-etal-2016-diversity} calculates the ratio between the number of unique n-grams and the total number of n-grams in a generation.

Table \ref{table:reframe_generate_results} shows the automatic evaluation results for the task. All the models perform close to each other in terms of BLEU, BERTScore, and ROUGE. GPT3.5 generates lexically diverse rewrites with the best Distinct-n scores. We provide examples of system outputs in Table \ref{table:system_outputs_main}.

\subsubsection{Human Evaluation}
As automatic metrics often fail to fully capture human preferences in text generation tasks, we also perform human evaluation. We collect human ratings  of 100 random thoughts from the test set. Similar to previous style transfer works \cite{ziems-etal-2022-inducing, briakou-etal-2021-review, rao-tetreault-2018-dear}, we evaluate the generated rewrites along three dimensions through Yes/No binary ratings: (i) fluency, which evaluates the readability of the generation, (ii) meaning preservation,  which here verifies if the rewrite aligns with the given persona and thought, and (iii)  quality, which here evaluates if the generated thought exhibits the given unhelpful thought pattern.   
We collect 9 annotations for each system output and apply majority voting to extract the final annotation.\footnote{We also provide results using a more stringent threshold of 7 out of 9 annotators rating positively, in Appendix \ref{app:results7}. The pattern of results is similar.}

Table \ref{fig:human_eval} shows the percentage of outputs rated positively by at least five of the nine annotators. GPT3.5 outperforms all other approaches, including human references, in terms of fluency and quality. However, GPT3.5 shows the lowest (but still very high) meaning preservation score for generating thoughts. The other models have more difficulty including the unhelpful pattern (lower "thought quality" scores). 

\begin{figure*}[th!]
    \begin{subfigure}{.5\textwidth}
      \centering
      \includegraphics[width=1.0\linewidth]{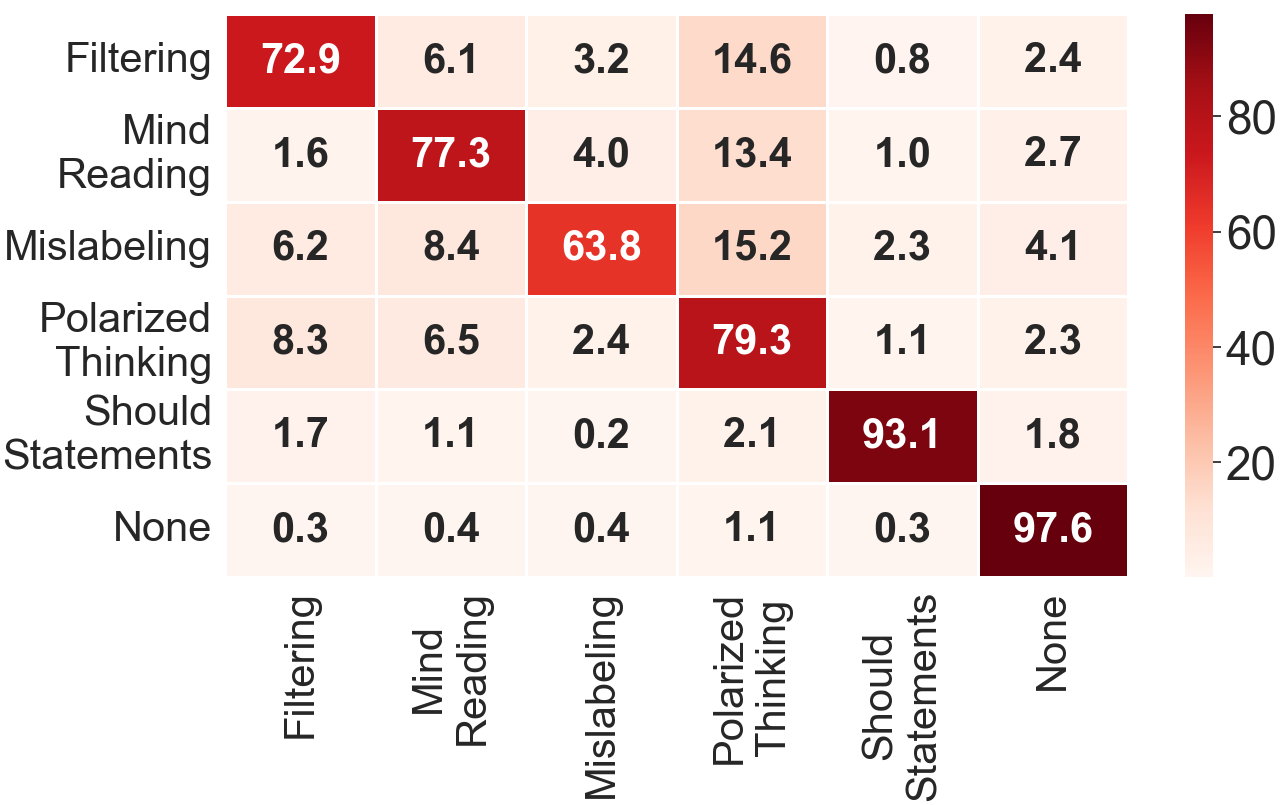}
      \caption{RoBERTa}
      \label{fig:sub1}
    \end{subfigure}
    \begin{subfigure}{.5\textwidth}
      \centering
      \includegraphics[width=1.0\linewidth]{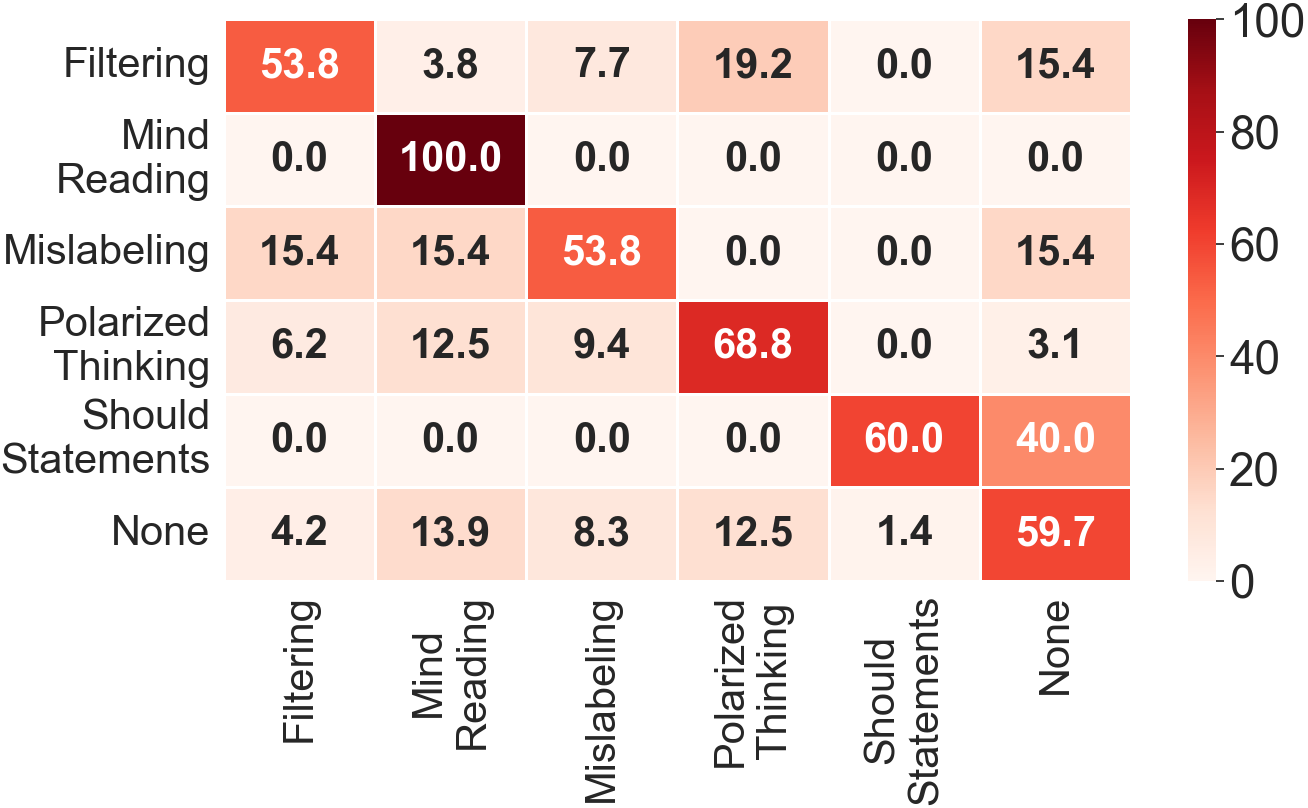}
      \caption{GPT3.5 (text-davinci-002)$\dagger$}
      \label{fig:sub1b}
    \end{subfigure}   
  \caption{Confusion matrices for the unhelpful thoughts classification task on our classification test set. The rows represent true labels and the columns represent predicted labels. We clustered similar patterns for clearer interpretation. \textit{Polarized Thinking} includes \textit{Overgeneralization}, \textit{Catastrophizing}, \textit{All or Nothing Thinking}, and \textit{Fortune Telling}. \textit{Filtering} refers to \textit{Mental Filtering} and \textit{Discounting the positive}. \textit{Mislabeling} encompasses \textit{Personalization} and \textit{Labeling and Mislabeling}.   $\dagger$We obtain outputs for only 100 random thoughts.}
  \label{fig:classification results}
\end{figure*}

\subsection {Classifying Unhelpful Thoughts}

\subsubsection{Task and Data}
Given a persona and a thought, the goal is to classify them into one of the ten unhelpful thought patterns or ``\textit{None}'', which indicates that the input thought does not contain any of the ten unhelpful patterns, or the thought does not align with the persona. We formulate the task as a multiclass classification problem with eleven categories. 

We once again use the same train, validation, and test splits described in \S \ref{sec:dataset_stats}. Note that the dataset contains only positive examples for the classification task, i.e., thoughts that align with a specific thought pattern and persona. For every positive example, we construct a negative example by randomly choosing one of the following options: (i) a thought from our dataset that belongs to the same pattern but a different persona. (ii) a dialog text from \textsc{Persona-Chat} belonging to the same persona (but presumably not containing any unhelpful pattern),  (iii) a dialog text from \textsc{Persona-Chat} belonging to a different persona (and again, presumably not containing any unhelpful pattern). Thus, negative examples encompass neutral texts and misaligned thoughts and personas. We assign the category ``None'' to these examples. We have 3,834 train, 1,915 validation, and 13,572 test instances after augmenting the dataset with these examples. 

\subsubsection{Methods}
 We finetune RoBERTa \cite{roberta2019} using the soft-label distribution obtained through the second task of our data collection pipeline (\S \ref{sec:annotation_pipeline}), 
 where we asked multiple annotators to identify the patterns exhibited in a thought, and then normalized the votes across the patterns. We use a soft label distribution instead of single label because of the high overlap across patterns. We also perform 11-way, 1-shot classification using GPT3.5. We construct the input prompt using one example from each category (examples in \S \ref{app:prompt_examples}) and classify 100 random inputs in the test set.  We include further implementation details in Appendix \ref{app:implementation details}.

\subsubsection{Evaluation}

Figure \ref{fig:classification results} shows the confusion matrices for RoBERTa and GPT3.5 on the augmented version of the \textsc{PatternReframe} test set. Given that several unhelpful thinking patterns are closely related (for example, All or Nothing Thinking and Catastrophizing), we cluster the patterns using the KMeans algorithm \cite{kmeans} to group together patterns that were deemed close by the model \footnote{We first constructed an aggregated vector for each pattern by averaging the 10-dimensional outputs of the classifier for the examples corresponding to each pattern on the validation set. We then clustered the ten 10-dimensional aggregated vectors into five clusters using the KMeans algorithm, with the number of clusters determined using the elbow method.}.  RoBERTa performs well on all the categories ($> 72\%$) except 
the \textit{Mislabeling} category, which has a high overlap with the \textit{Polarized Thinking} category. The \textit{None} category has the highest performance, which shows that the classifier is able to differentiate neutral texts that do not contain any unhelpful pattern, or texts that are not aligned with the persona. 1-shot classification using GPT3.5 performs worse than fine-tuned RoBERTa. GPT3.5 has trouble distinguishing texts with and without unhelpful patterns and gets a low score for \textit{None}. We also observed that 40\% of the classification predictions changed for GPT3.5 after reordering the sequence of examples in the prompt, which shows that few-shot classification is not as reliable for this task, while still providing decent performance way above chance.

\begin{table*}[ht!]
\centering
\small
\begin{tabular}{p{1.0cm}p{13.9cm}}
\toprule
 \multicolumn{2}{l}{\textbf{Generating Unhelpful Thoughts}} \\
\midrule
Persona &  I like to chat with my friends. I like to read books that are in a series. I like to go to the movies. I love to go out to eat with my family. \\
Pattern &  Catastrophizing  \\
T5 &  I love to go out to eat with my family, but I'm sure we'll all get sick and die. \\
BART &  I like to read books that are in a series, but I will never finish them. \\
R2C2 &  I love to go out to eat with my family, but one day we will all get food poisoning and die. \\
GPT3.5 &  I'm out of money to go out to eat with my family, maybe they won't want to hang out with me anymore. \\
Human &  My friends didn't want to chat with me today. They must have all hate me and wish me ill. \\
\midrule
Persona   &   I write children's books. I like to drink. I live alone. I shop online, too much. I don't believe in love. \\
Pattern   &   Should statements \\
T5   &  I write children's books, everyone should read them. \\
BART   &   I write children's books, everyone should read them. \\
R2C2   &   I write children's books. I should be published by now. \\
GPT3.5  &   I should be able to write the perfect book, drink moderately, be in a relationship, and not shop online so much. \\
Human   &   I shop online and I should make enough money to be able to afford anything and everything I want. \\
\midrule
 \multicolumn{2}{l}{\textbf{Reframing Unhelpful Thoughts}} \\
\midrule
Persona &  My mother was a teacher. My favorite food is a salad. I enjoy nature. I teach a yoga class. I am single. \\
Pattern &  Black-and-white or polarized thinking / All or nothing thinking \\
Thought &  The date I went on last week was a total fail. I'm going to be single forever. \\
T5 &  The date I went on last week was a total fail. I will try harder next time. \\
BART &  The date I went on last week was a total fail. Next time I will be more careful. \\
R2C2 &  The date I went on last week was a total fail. I'm going to be single forever. \\
GPT3.5 &  The date I went on last week wasn't a great match, but I'm open to meeting someone new and trying again. \\
Human &  I had a bad date last week, thankfully I went on a nature walk and realized I have plenty of time to find a partner. \\
\midrule
Persona &  I was diagnosed with mild OCD in college. Once a year, I take a week off and head to Florida. I freeze up in social situations. Sometimes I feel like I ve already been through life once before.\\
Pattern &  Personalization \\
Thought &  I was diagnosed with mild OCD in college, it's my fault and I deserve it. \\
T5 &  I was diagnosed with mild OCD in college. It's not my fault. \\
BART &  I was diagnosed with mild OCD in college. I am working hard to overcome it. \\
R2C2 &  I was diagnosed with mild OCD in college. I'm glad to have a diagnosis so I can get the help I need. \\
GPT3.5 &  I was diagnosed with mild OCD in college, it's something I'm learning to manage. \\
Human &  I was diagnosed with mild OCD in college. I've been seeing a therapist to get help managing it. \\
\bottomrule
\end{tabular}
\caption{Examples of system outputs for the tasks of generating and reframing unhelpful thoughts.}
\label{table:system_outputs_main}
\end{table*}

\subsection {Reframing Unhelpful Thoughts}

\subsubsection{Task and Methods}
\label{sec:reframing_task_and_data}
Given a persona, an unhelpful thought pattern, and a thought exhibiting the given pattern, the goal is to reframe the thought in a way that still aligns with the persona and the context of the thought but does not contain the pattern. The reframing problem is similar to the generating one, except that the unhelpful thought is now a part of the input instead of the target. We use the same training, validation, and test splits for the reframing task (\S \ref{sec:dataset_stats}). We also evaluate the same approaches described in \S \ref{sec:generation_methods}. For fine-tuned methods, we concatenate the persona, the pattern, and the thought texts with a special token. For few-shot methods, we construct a prompt similar to the one used for the generation task, but with the reframed statements (examples in Appendix \ref{app:prompt_examples}).

\subsubsection{Automatic Evaluation}

Table \ref{table:reframe_generate_results} shows the automatic evaluation results on the \textsc{PatternReframe} test set.  We use the metrics described in \S \ref{sec:reframing_automatic_metrics} namely BLEU, ROUGE, BERTScore, and Distinct-n metrics. As each unhelpful thought can have up to 3 ground truth reframed versions, we take the maximum of the three scores and report the mean of these maxima. R2C2 performs the best in terms of BLEU and BERTScore.  GPT3.5 again outperforms the other models and the human references in terms of Distinct-1 and Distinct-2 scores, which indicates that the generations are lexically diverse. Table \ref{table:system_outputs_main} provides examples of system outputs.

\subsubsection{Human Evaluation}

Figure \ref{fig:human_eval} shows human evaluation results on 100 reframed thoughts generated by different models given the persona, the pattern type, and the unhelpful thought from our test set. Similar to the generating thoughts task, we evaluate the reframed thoughts along fluency, meaning preservation, and quality, where we ask the annotators if the reframed thought removes the given unhelpful pattern while being consistent with the initial thought. All models perform close to human reference in terms of fluency and meaning preservation. In fact, all the outputs of R2C2 and GPT3.5 are fluent and preserve meaning (that is, they generate statements that are not contradictory with the initial thought). For reframing quality, that is, removing the unhelpful pattern, all models perform over 70\%, but GPT3.5 performs the best. GPT3.5's superiority is even more marked when using the more stringent threshold of 7 out of 9 annotators rating positively in Appendix \ref{app:results7}.

Overall, the evaluation suggests that using modern models to produce reframing is a feasible approach, even with a small amount of data for fine-tuning. In particular, GPT3.5 performs remarkably well and very close to crowdsource worker performance, only based on prompting.

\section{Conclusion}
In this work, we introduced a novel dataset, \textsc{PatternReframe}, which contains (1) about 10k statements exhibiting unhelpful thought patterns, conditioned on a persona, and (2) multiple rewritten complementary thoughts that do not contain the initial unhelpful pattern, instead reframing the thought in a more constructive way.

Using this dataset to train or prompt various modern language models, we showed that this range of models can already be a powerful tool to generate, identify, and reframe unhelpful thoughts, conditioned on a persona. 
By releasing our dataset \footnote{\url{https://github.com/facebookresearch/ParlAI/tree/main/projects/reframe_thoughts}}, we hope to help practitioners of CBT draw from a richer, more diverse set of examples of unhelpful thought patterns and reframings. This would help address the important limitation of a lack of personalized and specific examples in existing datasets, when teaching cognitive techniques.

Future work will evaluate whether leveraging models to produce richer training material results in more robust learning and understanding of the types of unhelpful thought patterns in humans.This may serve as the basis for future psychological validation studies of the materials and support future studies of low-intensity self-help interventions.

\section{Limitations}

This work relied on previously published datasets to source personas on which to anchor the generated unhelpful thoughts, and thus shares the limitations of those datasets. In particular, they use English-language responses, written by workers located in the United States.\footnote{Our crowdsourcing tasks pay workers well above minimum wage. 
}. While these workers are reasonably diverse \citep{moss2020demographic}, the examples generated may not reflect the thought patterns and personas across cultures and diverse populations. This data is also generated by people who are being paid, as opposed to people genuinely engaging about situations that matter to them. Besides the substance of the thoughts themselves, a more direct limitation is that the models generate only English, so would not be directly usable for speakers of other languages.

In addition, the data collected reflects the understanding of lay people, rather than trained clinical psychologists. While this makes the material more immediately relatable to other lay people, it is possible that the data do not capture what clinical psychologists would consider adequate illustrations of unhelpful patterns. Our data has been spot-checked by a CBT-trained clinical psychologist and found generally sound, but the entire material should undergo further validation.

Another limitation is that the models that we have tested are resource-intensive. In particular, the best-performing model, GPT3.5, is only available through a paid API. 

\section{Ethical considerations}

While our work was developed to generate abundant data supporting work towards improving well-being, the negative statements it generates could be misused. The parallel data of unhelpful thoughts and their reframed versions can also be used to generate negative texts from neutral ones, by training systems with reframed versions as the input and unhelpful thoughts as the output. This risk of generating negative content from positive/neutral texts aligns with the risks of toxicity reduction and sentiment style transfer tasks.

Conversely, a different risk stems from over-eager use of our work. This work aims to examine the feasibility of generating ample practice material anchored on specific personas. We hope that releasing a large dataset of unhelpful thoughts and reframings will further research that will ultimately help practitioners, but there is a danger that people attempt to use the material as is, without the supervision of a trained professional, which could be harmful, as the material has not been tested with participants while monitoring adverse events such as increased anxiety or warped understanding of what unhelpful thoughts and useful reframings are.

\bibliography{anthology,custom}

\appendix

\newpage
\onecolumn
\section{Data Collection Details}
\label{app:annDetails}

\subsection{Onboarding Tasks}

We introduce two onboarding tasks to ensure that the crowdsource workers understood the concept of unhelpful thoughts and how to reframe them. The onboarding tasks were reviewed by a CBT-trained psychologist. We use one onboarding task for tasks 1 and 2 and another onboarding task for tasks 3 and 4 of the data collection pipeline. For the first onboarding task, we display an unhelpful thought pattern, one positive example that contains the pattern, and one negative example that does not, and ask the workers to select the positive one. We only allowed the workers that were able to identify the correct example for three out of four such instances. For the second onboarding task,  we display  an unhelpful thought pattern, a thought containing the pattern, one positive example that reframes the thought, and one negative example that does not. We only allow the workers that were able to identity the positive example in three out of four such instances. 

\section{Data Collection Interface Snapshots}
\label{app:ann_interface}

\begin{figure}[h]
    \centering
    \includegraphics[width=0.92\textwidth]{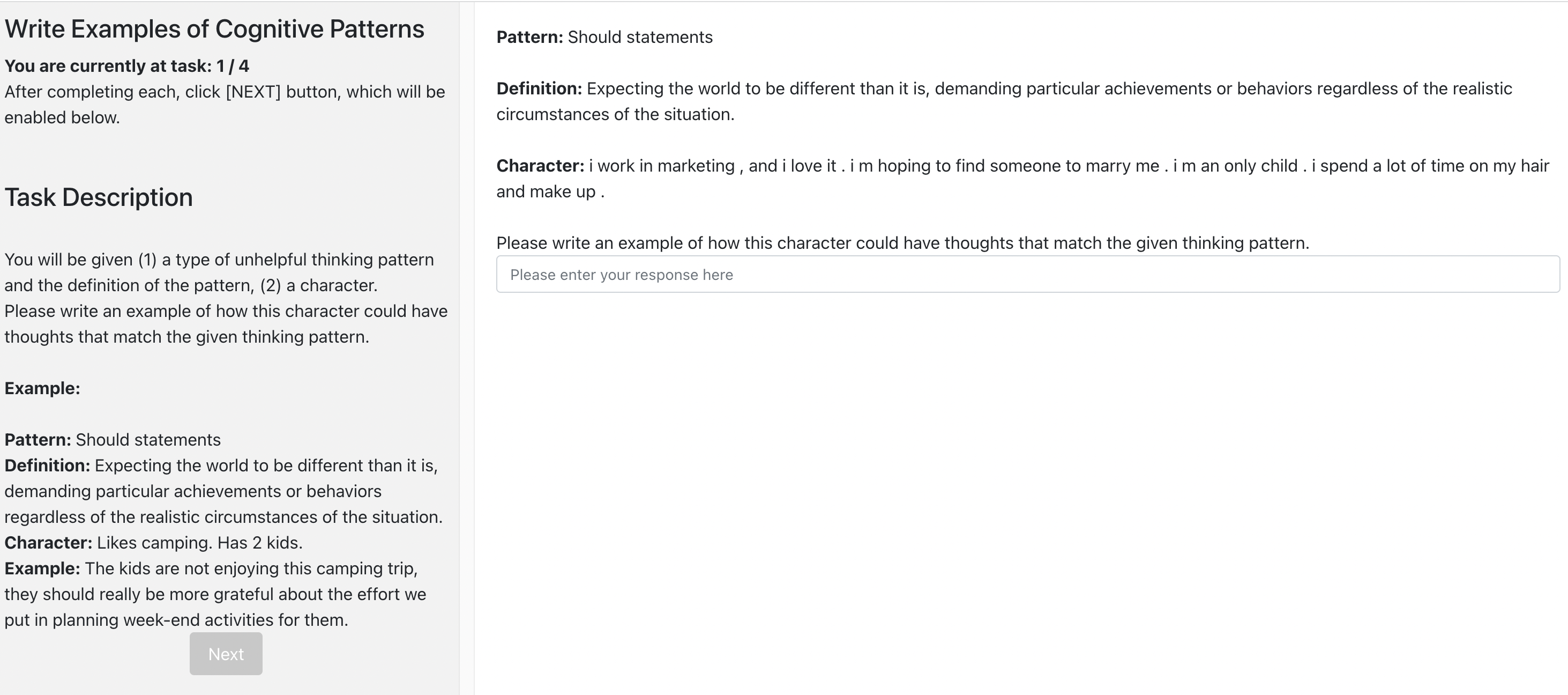}
    \captionof{figure}{Data collection interface for the first task of the data collection pipeline, where crowdworkers are asked to write an unhelpful thought.}
    \label{fig:task1}
\end{figure}

\begin{figure}[h]
    \centering
    \includegraphics[width=0.92\textwidth]{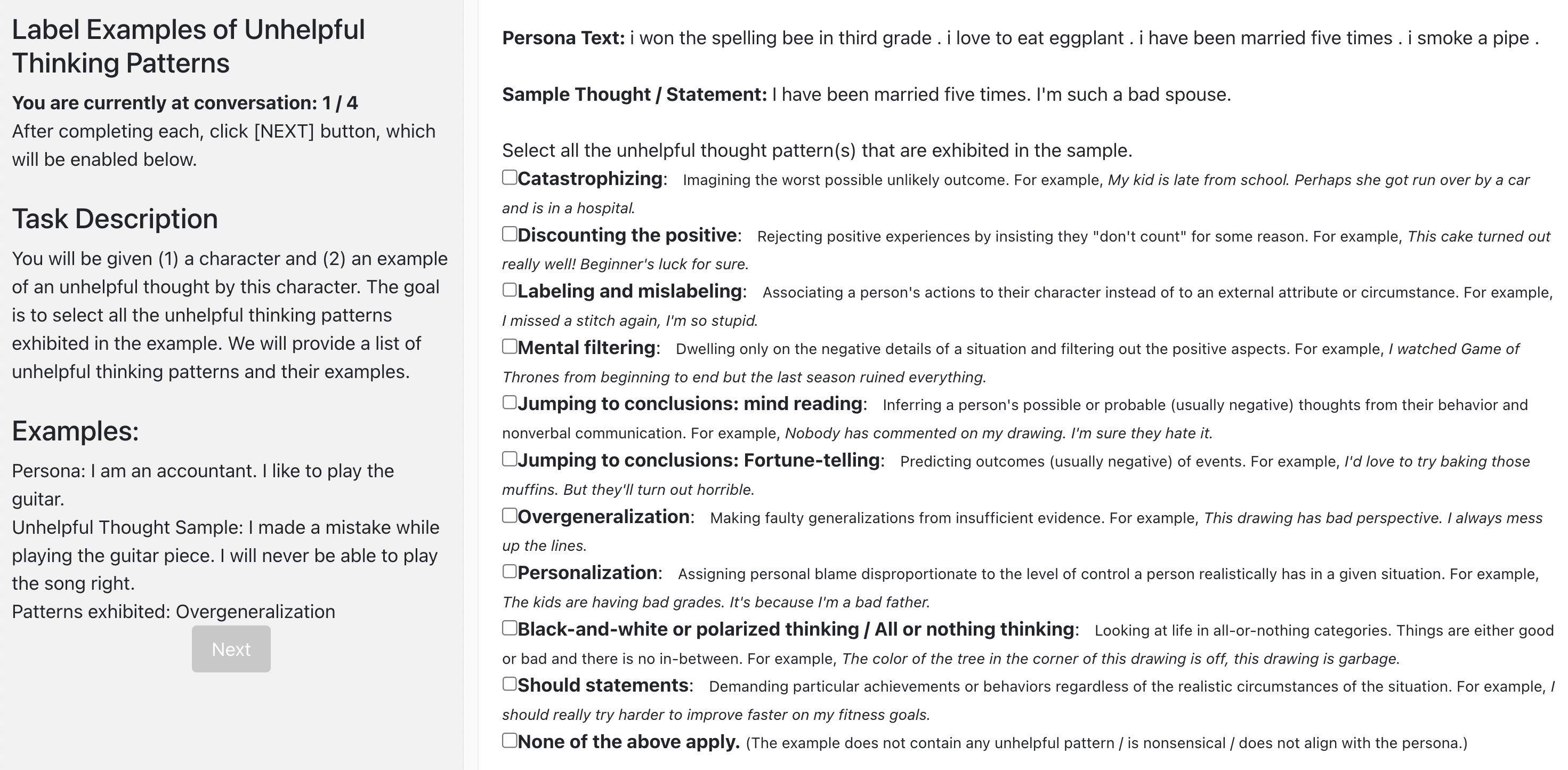}
    \captionof{figure}{Annotation interface for the second task of the data collection pipeline, where crowdworkers are asked to select the patterns exhibited by an unhelpful thought.}
    \label{fig:task2}
\end{figure}

\clearpage

\begin{figure}[h]
    \centering
    \includegraphics[width=\textwidth]{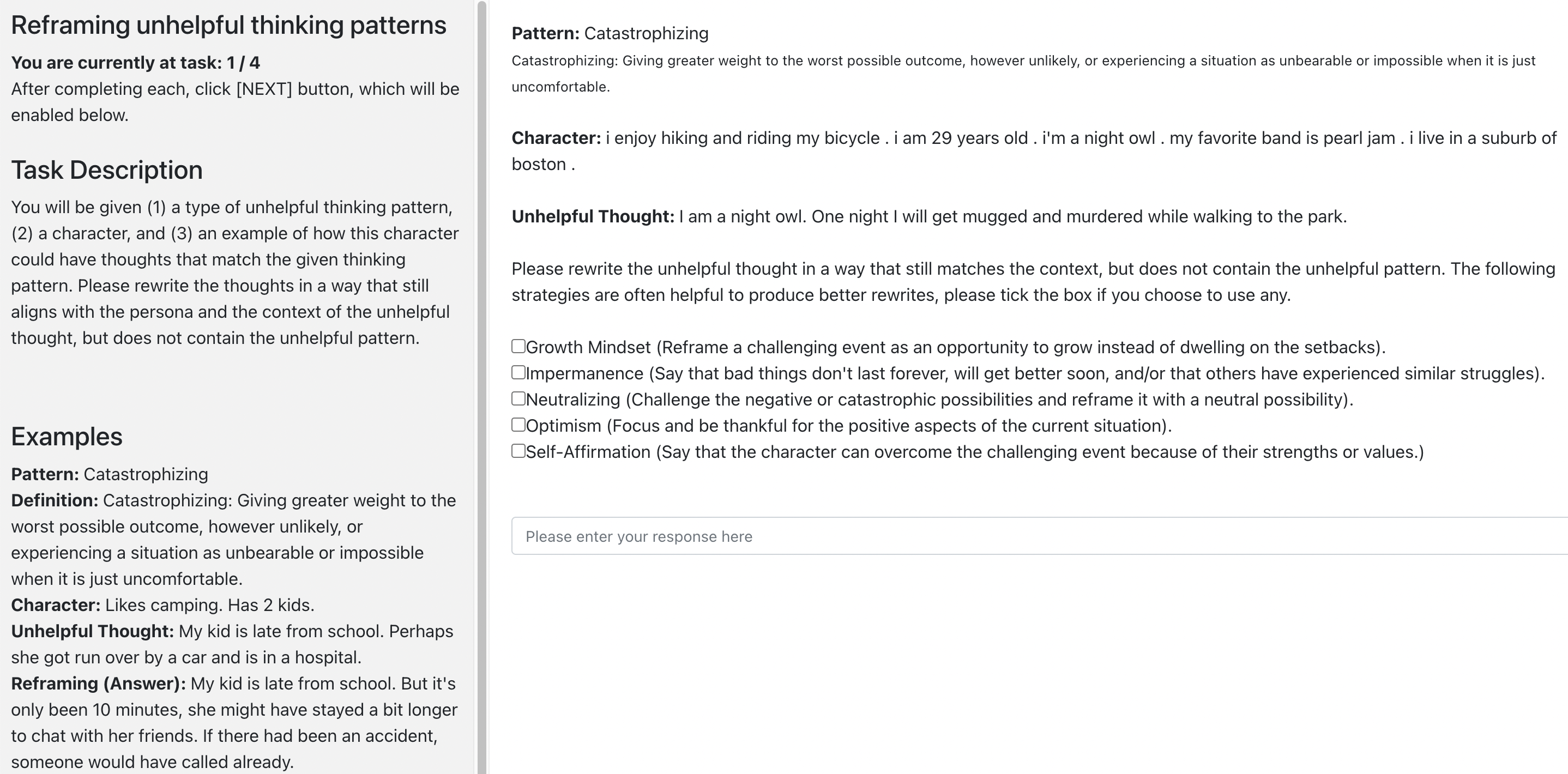}
    \captionof{figure}{Data collection interface for the third task of the data collection pipeline, where the crowdworkers are asked to reframe unhelpful thoughts.}
    \label{fig:task3}
\end{figure}

\begin{figure}[h]
    \centering
    \includegraphics[width=\textwidth]{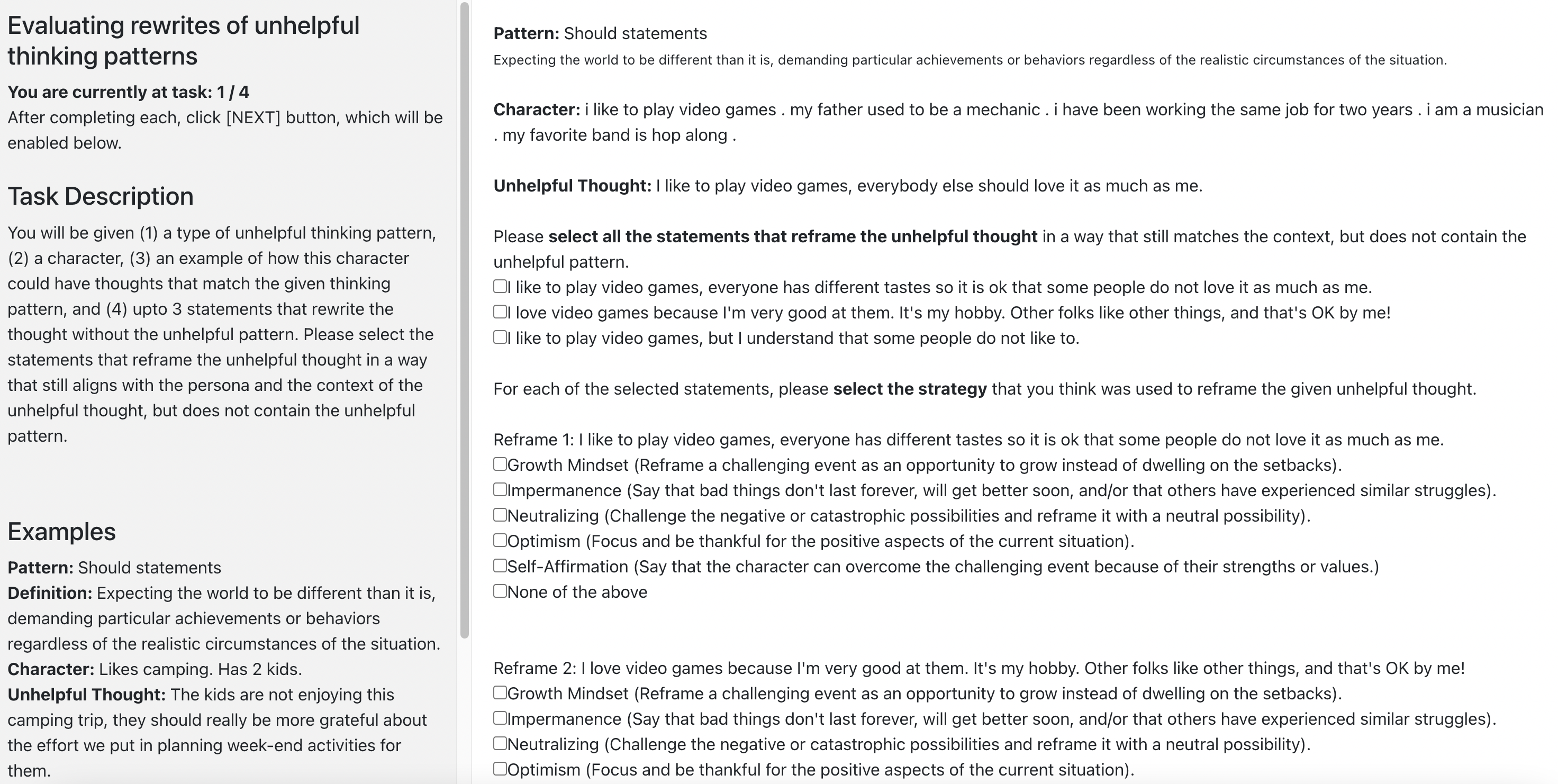}
    \captionof{figure}{Annotation interface for the fourth task of the data collection pipeline, where the crowdworkers are asked to evaluate the quality of the reframed thoughts.}
    \label{fig:task4}
\end{figure}

\clearpage
\section{Evaluation Interface Snapshots}
\label{app:eval_interface}

\begin{figure}[h]
    \centering
    \includegraphics[width=0.6\textwidth]{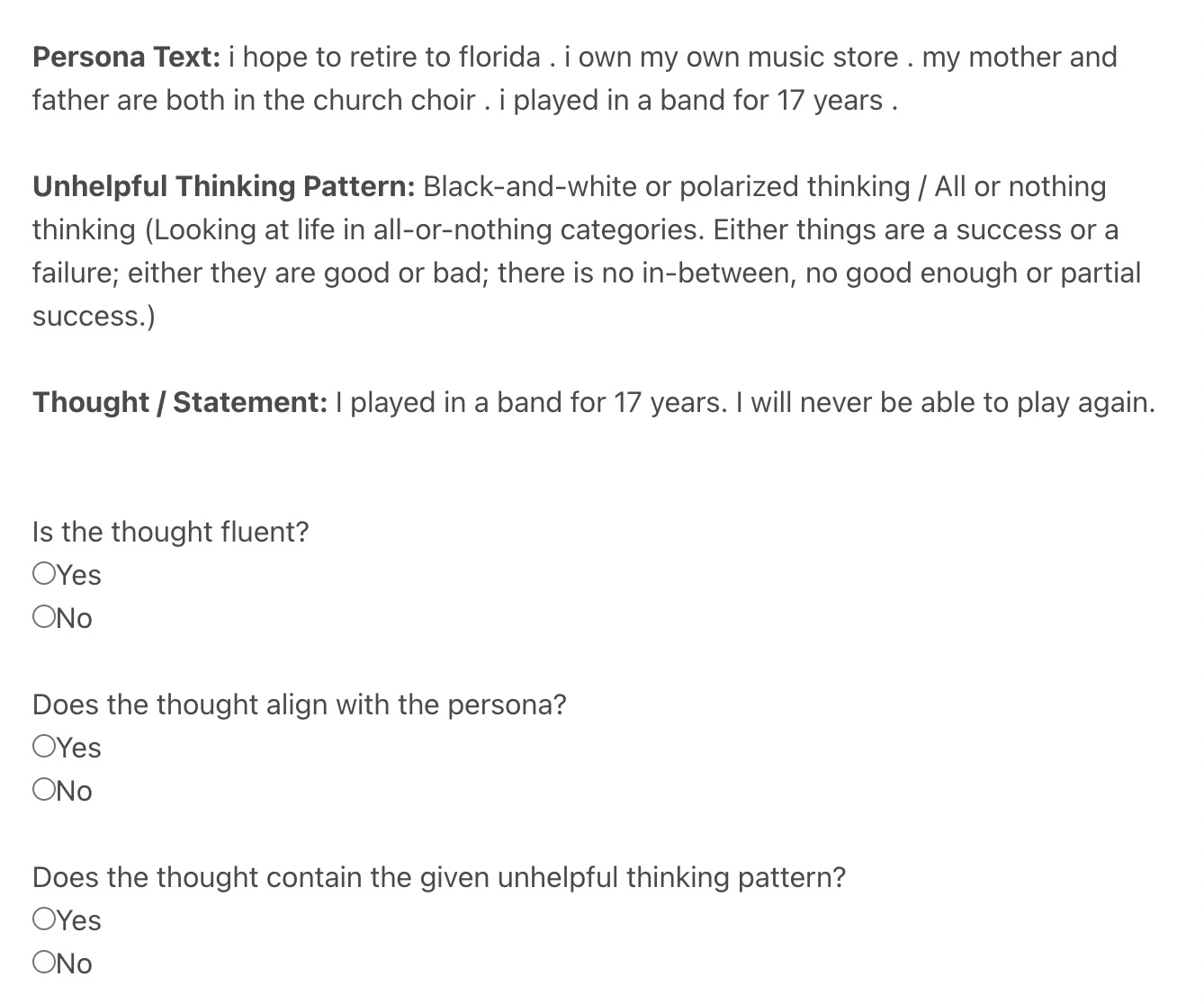}
    \captionof{figure}{Annotation interface used to evaluate generated thoughts.}
    \label{fig:generate_eval}
\end{figure}

\begin{figure}[h]
    \centering
    \includegraphics[width=0.6\textwidth]{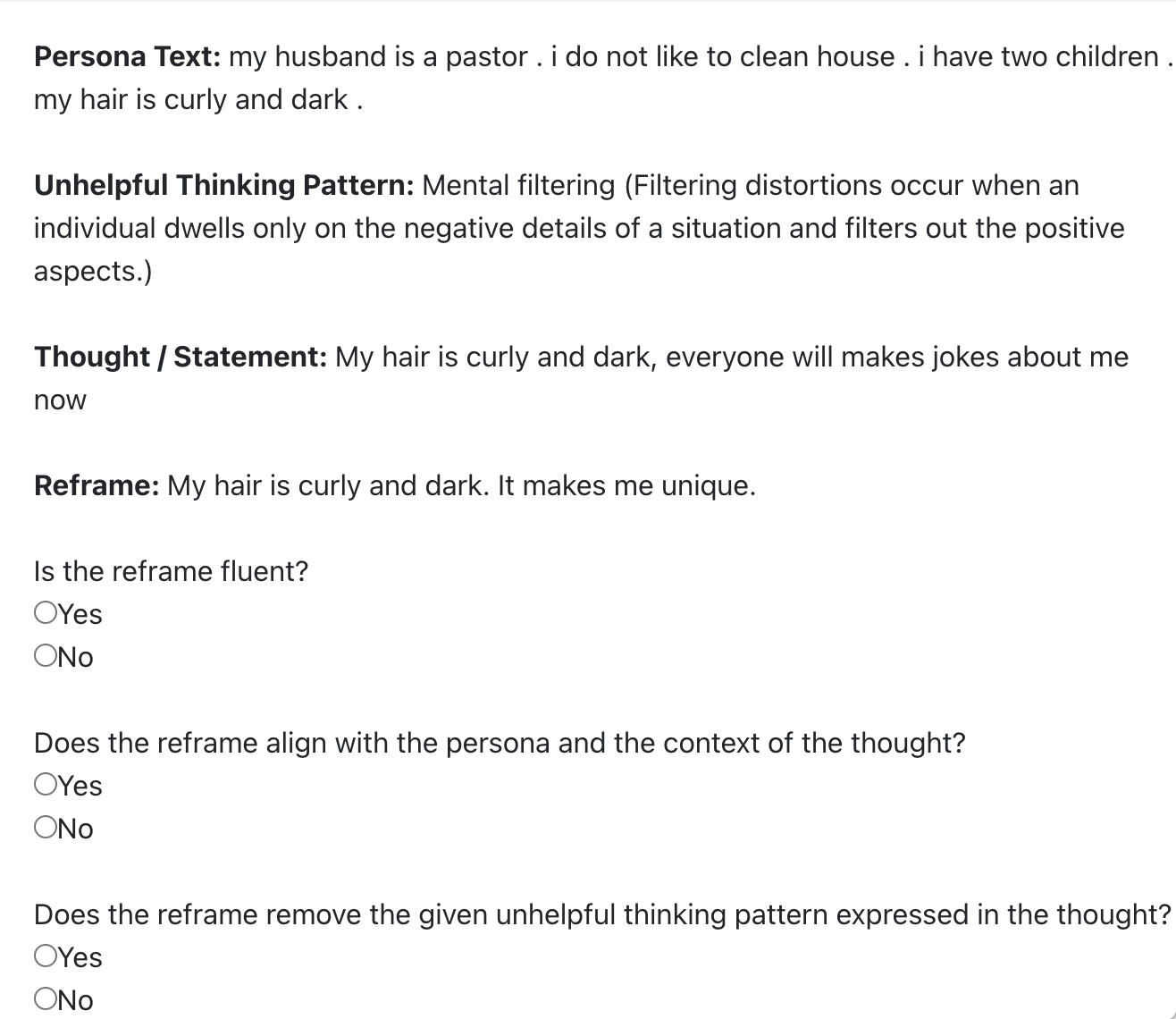}
    \captionof{figure}{Annotation interface used to evaluate statements that reframe unhelpful thoughts.}
    \label{fig:reframe_eval}
\end{figure}

\clearpage

\section{Implementation details}
\label{app:implementation details}

\subsection{Generation Models}
We finetuned the BART, T5, and R2C2 baselines using ParlAI\footnote{https://www.parl.ai/docs/index.html}. We used the BART$_{large}$ (400M parameters), T5$_{large}$ (770M parameters), and R2C2$_{base}$ (2.7b parameters)\footnote{https://parl.ai/docs/zoo.html\#r2c2-base-2-7b} architectures. We used Adam optimizer \cite{Kingma2014} and performed a hyperparameter search over learning rates 1e-05, 1e-06, 1e-07, and 1e-08. We used linear warmup of 100 steps and applied early stopping with a patience value of 5. We evaluated the validation set once in every 200 updates and truncated the input and the labels to 1000 tokens. We applied gradient clipping value of 1.0. We used a batch size of 32. During inference, we used beam search with beam size 10. We chose the best checkpoint during training based on the perplexity on the validation set. Each model takes around 1 hour to run on 8 NVIDIA Tesla V100 Volta 32GB GPUs.

\subsection{Classification Models}
For classification experiments, we finetuned the RoBERTa-large checkpoint from Huggingface\footnote{https://github.com/huggingface/transformers}.  We used Adam optimizer \cite{Kingma2014}, learning rate of 1e-05, with linear warmup of 100 steps. We trained the model for a maximum of 10 epochs. We evaluated on the validation set every 200 updates.  We used a batch size of 16. We chose the best checkpoint during training based on the weighted F1 value on the validation set. The model takes around 1 hour to run on 1 NVIDIA Tesla V100 Volta 32GB GPU.

\section{GPT3.5 Prompt Examples}
\label{app:prompt_examples}

\begin{table*}[ht!]
\small
\centering
\begin{tabular}{p{15cm}}
\toprule
You will be given (1) a type of unhelpful thinking pattern and the definition of the pattern and (2) a character. Please write an example of how this character could have thoughts that match the given thinking pattern. 
\newline
\newline
Persona: Likes camping. Has 2 kids. 
\newline
Unhelpful Thinking Pattern: Discounting the positive (Rejecting positive experiences by insisting they "don't count" for some reason or other.)
\newline
Unhelpful Thought: My friends said they really enjoyed the camping trip I organized, but anyone could have done it.
\newline
\newline
Persona: i'm a business man. i love to sing. i'm a karate black belt. my wife has terminal cancer. 
\newline
Unhelpful Thinking Pattern: Discounting the positive (Rejecting positive experiences by insisting they "don't count" for some reason or other.)
\newline
Unhelpful Thought: \\
\bottomrule 
\end{tabular}
\caption{Example GPT3.5 prompt for the task of generating unhelpful thoughts.}
\label{table:generate_thought_prompt_example}
\end{table*}


\begin{table*}[h!]
\small
\centering
\begin{tabular}{p{15cm}}
\toprule
You will be given a type of unhelpful thinking pattern, a character, and an example of how this character could have thoughts that match the given thinking pattern. Please rewrite the thoughts in a way that still aligns with the persona and the context of the unhelpful thought, but does not contain the unhelpful pattern.
\newline
\newline
Persona: Likes camping. Has 2 kids. 
\newline
Unhelpful Thinking Pattern: Overgeneralization (Someone who overgeneralizes makes faulty generalizations from insufficient evidence. Even if something bad happens only once, it is expected to happen over and over again.)
\newline
Unhelpful Thought: My younger kid has gotten bad grades at his maths test this week. He'll never be good at maths.
\newline
Reframe: My younger kid has gotten bad grades at his maths test this week. It's been a few times but hopefully we can figure out a way to help him get better.
\newline
\newline
Persona: i obsess over working out and being the best . i got a scholarship for playing soccer . its important for my instagram posts to look like i am having fun . i try to eat healthy or i don't eat at all .
\newline
Unhelpful Thinking Pattern: Overgeneralization (Someone who overgeneralizes makes faulty generalizations from insufficient evidence. Even if something bad happens only once, it is expected to happen over and over again.)
\newline
Unhelpful Thought: My future college team lost another game, I will never become a good athlete playing for them.
\newline
Reframe: \\
\bottomrule
\end{tabular}
\caption{Example GPT3.5 prompt for the task of reframing unhelpful thoughts.}
\label{table:reframe_thought_prompt_example}
\end{table*}

\newpage

\begin{table*}[h!]
\small
\centering
\begin{tabular}{p{15cm}}
\toprule
Persona: Likes camping. Has 2 kids. \newline
Unhelpful Thought: The kids have stopped paying attention to how we can pitch the tent. They will never learn. \newline
Unhelpful Thinking Pattern: Jumping to conclusions: Fortune-telling \newline
\newline
Persona: Likes camping. Has 2 kids. \newline
Unhelpful Thought: The kids are not enjoying this camping trip, they should really be more grateful about the effort we put in planning week-end activities for them. \newline
Unhelpful Thinking Pattern: Should statements \newline
\newline
Persona: Likes camping. Has 2 kids.  \newline
Unhelpful Thought: My kid is late from school. Perhaps she got run over by a car and is in a hospital. \newline
Unhelpful Thinking Pattern: Catastrophizing \newline
\newline
Persona: Likes camping. Has 2 kids. \newline
Unhelpful Thought: This camping trip was a catastrophe. Sure the weather was gorgeous and the kids had a lot of fun, but the waterfall always had many people ruining the photos we wanted to take. \newline
Unhelpful Thinking Pattern: Mental filtering \newline
\newline
Persona: Likes camping. Has 2 kids.  \newline
Unhelpful Thought: I like camping with my kids. We had a lot of fun the other weekend.\newline
Unhelpful Thinking Pattern: None\newline
\newline
Persona: Likes camping. Has 2 kids. 
Unhelpful Thought: The kids are having bad grades. It's because I'm a bad father. \newline
Unhelpful Thinking Pattern: Personalization \newline
\newline
Persona: Likes camping. Has 2 kids. \newline
Unhelpful Thought: My younger kid has gotten bad grades at his math test this week. He'll never be good at math. \newline
Unhelpful Thinking Pattern: Overgeneralization \newline
\newline
Persona: Likes camping. Has 2 kids. \newline
Unhelpful Thought: My friends said they really enjoyed the camping trip I organized, but anyone could have done it. \newline
Unhelpful Thinking Pattern: Discounting the positive \newline
\newline
Persona: Likes camping. Has 2 kids. \newline
Unhelpful Thought: My kids are being very silent. I am sure it's because they really hate me for taking them on this camping trip. \newline
Unhelpful Thinking Pattern: Jumping to conclusions: mind reading\newline
\newline
Persona: Likes camping. Has 2 kids.  \newline
Unhelpful Thought: I didn't manage to light up the fire for the camp today, I'm such a useless outdoors person. \newline
Unhelpful Thinking Pattern: Labeling and mislabeling \newline

Persona: Likes camping. Has 2 kids. \newline
Unhelpful Thought: One of the 5 trails we planned to do on this trip is closed to the public. This trip is ruined. \newline
Unhelpful Thinking Pattern: Black-and-white or polarized thinking / All or nothing thinking \newline
\newline
Persona: i'm a woman . i've several children . we have a dog . we live in a rural area . my parents are still married . \newline
Unhelpful Thought: congratulations ! have you graduated college ? i am attending the university of michigan in the fall . \newline
Unhelpful Thinking Pattern: \newline \\
\bottomrule
\end{tabular}
\caption{Example GPT3.5 prompt for the task of identifying unhelpful thoughts.}
\label{table:classify_thought_prompt_example}
\end{table*}

\clearpage
\newpage
\section{Results with 7 over 9 agreement}
\label{app:results7}

\begin{figure}[h]
    \centering
    \includegraphics[width=\textwidth]{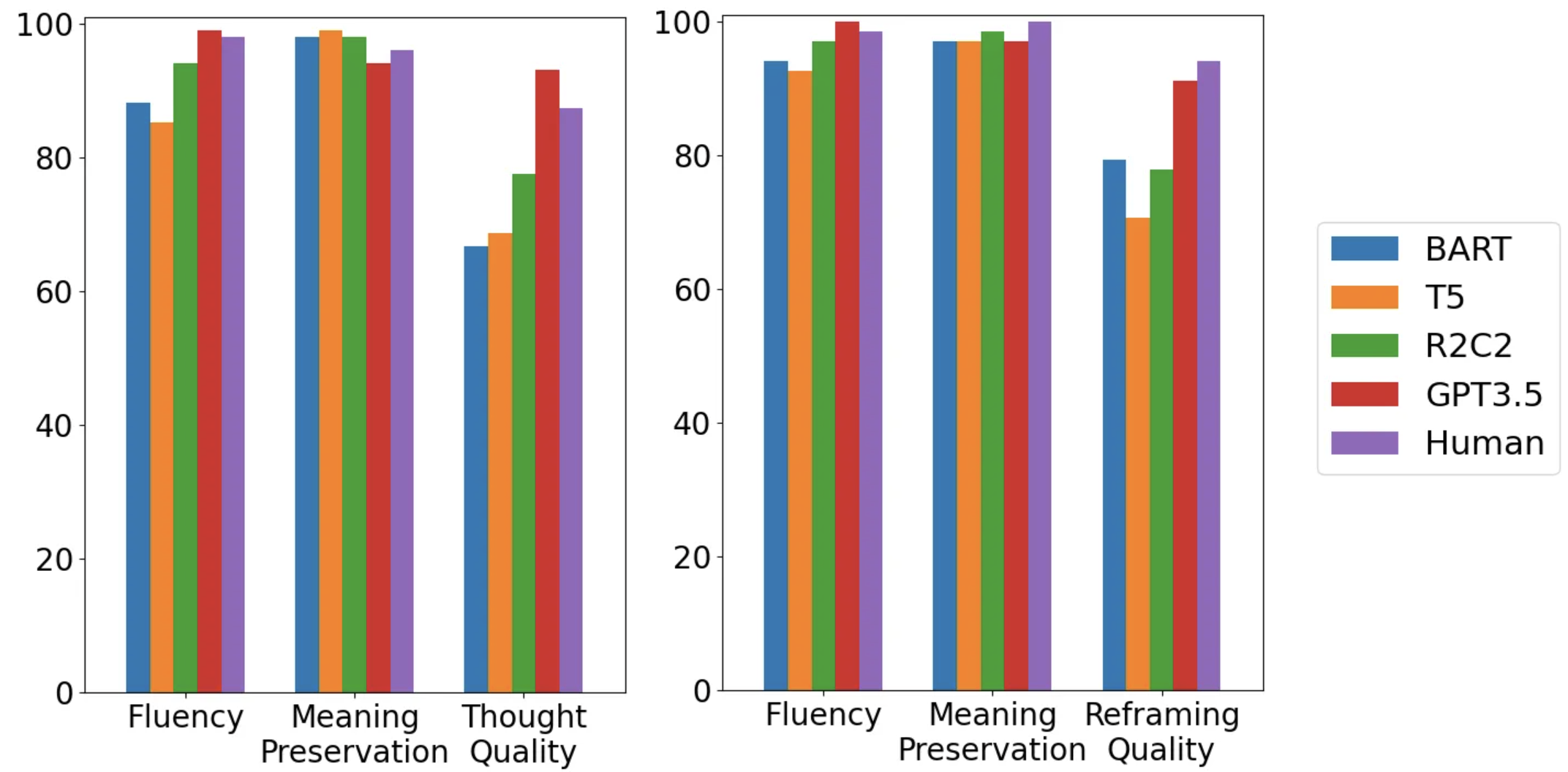}
      \caption{Human evaluation results for the tasks of  generating (left) and reframing  (right) unhelpful thoughts. Y-axis shows the percentage of outputs rated positively by at least seven of the nine annotators.}
     \label{fig:human_eval7}
\end{figure}

\end{document}